\documentclass[10pt,twocolumn,letterpaper]{article}

\usepackage[final]{cvpr}      % To produce the REVIEW version
\usepackage{multirow}
\usepackage{graphicx}
\usepackage{subcaption}

\definecolor{cvprblue}{rgb}{0.21,0.49,0.74}
\usepackage[pagebackref,breaklinks,colorlinks,allcolors=cvprblue]{hyperref}

\def\paperID{*****} % *** Enter the Paper ID here
\def\confName{CVPR}
\def\confYear{2026}

\title{API: Empowering Generalizable Real-World Image Dehazing via Adaptive Patch Importance Learning}

\author{
Chen Zhu\quad
Huiwen Zhang\quad
Yujie Li\quad 
Mu He\quad 
Xiaotian Qiao\footnotemark[2]
\\
\\
 Xidian University
}

\begin{document}
\maketitle

\begin{abstract}
Real-world image dehazing is a fundamental yet challenging task in low-level vision.
Existing learning-based methods often suffer from significant performance degradation when applied to complex real-world hazy scenes, primarily due to limited training data and the intrinsic complexity of haze density distributions.
To address these challenges, we introduce a novel Adaptive Patch Importance-aware (API) framework for generalizable real-world image dehazing.
Specifically, our framework consists of an Automatic Haze Generation (AHG) module and a Density-aware Haze Removal (DHR) module.
AHG provides a hybrid data augmentation strategy by generating realistic and diverse hazy images as additional high-quality training data.
DHR considers hazy regions with varying haze density distributions for generalizable real-world image dehazing in an adaptive patch importance-aware manner.
To alleviate the ambiguity of the dehazed image details, we further introduce a new Multi-Negative Contrastive Dehazing (MNCD) loss, which fully utilizes information from multiple negative samples across both spatial and frequency domains. 
Extensive experiments demonstrate that our framework achieves state-of-the-art performance across multiple real-world benchmarks, delivering strong results in both quantitative metrics and qualitative visual quality, and exhibiting robust generalization across diverse haze distributions.
\end{abstract}

\section{Introduction}
\label{sec:intro}
Haze is a common atmospheric phenomenon triggered by the presence of small airborne particles. It significantly reduces image visibility through low contrast and color distortion, severely impacting vision tasks such as scene understanding \cite{scene_understanding} and object detection \cite{detection}.

\begin{figure*}[t]
\centering
\includegraphics[width=\linewidth]{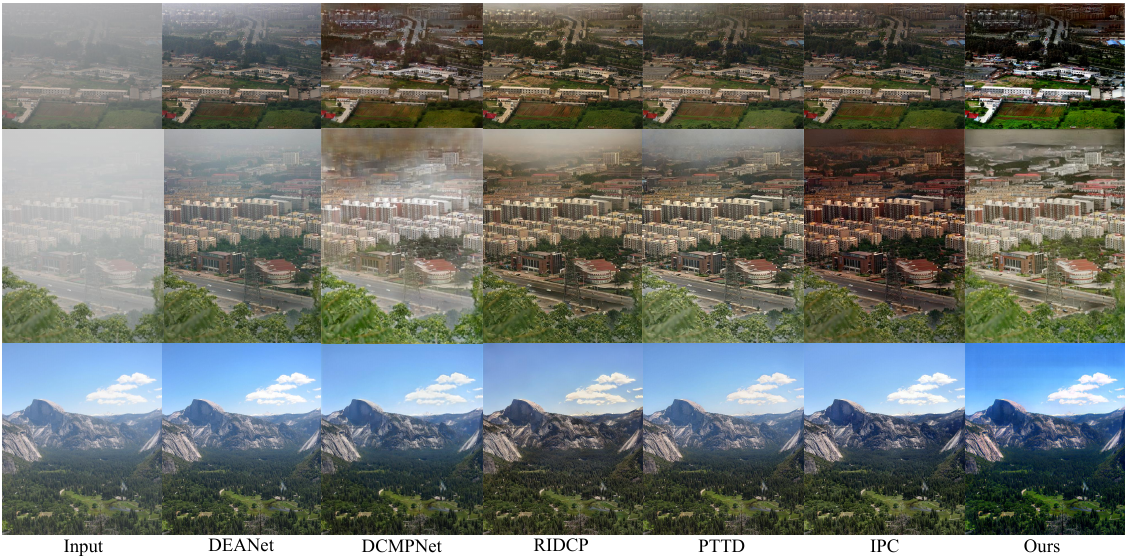} 
\caption{Visual comparison of dehazing results on real-world images. Unlike synthetic data, haze in real-world scenes typically exhibits diverse distributions, including dense haze (first row), non-uniform haze (second row), and light haze (third row). Our method produces more realistic and comprehensive dehazing results than previous state-of-the-art approaches.}
\label{fig:intro}
\end{figure*}

The objective of the single-image dehazing \cite{survey2,survey5,Cycle-dehaze,fusion, bmvc4, bmvc5} is to restore the visual information from an observed hazy image. In recent years, image dehazing has witnessed substantial progress, particularly driven by deep learning-based methods\cite{FAMED-Net,Ri-gan,bmvc3}.

However, obtaining large-scale, high-quality paired real-wrold hazy and clear images is higly impractical. Consequently, most methods either depend on synthetic datasets \cite{SOTS} or use limited real-world datasets \cite{NH-HAZE}, typically containing only 30–40 image pairs. This scarcity, combined with the significant domain gap between synthetic and real hazy images, hampers generalization and results in degraded performance under complex real-world conditions.

Moreover, as illustrated in Figure~\ref{fig:intro}, real-world haze distributions are typically highly complex, characterized by substantial variations in both density and spatial structure, including dense (first row), non-uniform (second row), and light haze (third row). Such variations present significant challenges for achieving consistent restoration across haze distributions with diverse characteristics.
As a result, many existing methods, particularly those trained on synthetic or homogeneous datasets, struggle to generalize well and often produce suboptimal results when applied to diverse real-world scenarios.

In light of these challenges, we are interested in a fundamental question: \textit{Is there a generalizable dehazing method capable of handling real-world hazy images with diverse and complex haze distributions?}  
Taking this into consideration, we propose \textit{\textbf{API}}, an \textbf{A}daptive \textbf{P}atch \textbf{I}mportance learning paradigm designed for real-world image dehazing.  
\textit{API} comprises two key components: an \textbf{A}utomatic \textbf{H}aze \textbf{G}eneration (AHG) module and a \textbf{D}ensity-aware \textbf{H}aze \textbf{R}emoval (DHR) module.

Firstly, to address the scarcity of real-world hazy datasets, we propose AHG as a hybrid data augmentation strategy. AHG encodes real images into haze density maps, which are then decoded to reconstruct realistic hazy images. By weighting different density maps to simulate varying haze concentrations and perturbing them with Perlin noise to model regional haze variations, AHG effectively generates diverse hazy images with distinct spatial haze patterns. This strategy significantly enriches the training data, thereby enhancing the model’s generalization to complex real-world hazy scenes and mitigating the domain gap between synthetic and real-world haze distributions.

We further propose DHR to effectively address spatially diverse haze distributions. Built on a U-Net architecture, DHR performs patch-level enhancement on hazy images both before and after passing through the U-Net backbone. Specifically, it partitions the input image into patches and extracts discriminative spatial and frequency features independently. Adaptive residual connections are employed to contextually modulate the processing of each patch based on its local haze density and background content, significantly improving the network’s ability to handle complex haze patterns. After patch-wise processing, all patches are seamlessly recombined to ensure global consistency and suppress boundary artifacts.

Additionally, we introduce the Multi-Negative Contrastive Dehazing (MNCD) loss, which leverages the hazy images generated by AHG as multiple negative samples in the contrastive loss to guide effective haze removal across varying haze densities.

We conduct extensive experiments on four real-world paired hazy datasets, as well as three real-world unpaired hazy dataset.  
Experimental results show that our model consistently surpassing existing methods in both quantitative evaluation and qualitative perception.  

To summarize, we make the first attempt to explore adaptive patch importance learning and propose a novel paradigm for real-world image dehazing. To improve generalization and mitigate domain gap, we introduce a hybrid data augmentation strategy and a patch-wise density-aware haze removal approach to effectively handle the complex and diverse haze distributions in real-world scenarios. Extensive experiments on multiple real-world benchmarks demonstrate that our method consistently achieves state-of-the-art performance.

\section{Related Work}
\paragraph{Prior Based Dehazing Methods.}
Early dehazing methods employ the Atmospheric Scattering Model (ASM) \cite{TASM} to describe the formation of hazy images, aiming to estimate its parameters for image restoration. To constrain this ill-posed problem, various physical priors have been proposed, including the Dark Channel Prior (DCP) \cite{DCP}, the color lines prior \cite{fattal}, the color attenuation prior \cite{color_attenuation}, and the maximum reflectance prior \cite{maximum_prior}.
More recent approaches seek to incorporate prior knowledge into learning frameworks through the design of loss functions or network architectures.
For instance, PSD \cite{PSD} proposes a prior loss committee to fine-tune the network on real-world data in an unsupervised manner.
RIDCP \cite{RIDCP} and IPC \cite{IPC} incorporate a codebook prior into the dehazing network, aiming to better encode and utilize haze-related features. 
Despite these advancements, most of the above methods are fundamentally reliant on predefined physical priors which may not generalize well to complex real-world scenes.
\paragraph{Deep Learning Based Dehazing Methods.}
Early deep learning based approaches \cite{dehazenet,aod,MSCNN,bmvc2} employ convolutional neural networks (CNNs) to estimate the parameters of the degradation model defined by the Atmospheric Scattering Model (ASM).
Subsequent works \cite{PMNet,dehaze,dehaze2,dehaze3,dehaze4,BMVC,Defog_BMVC} shift towards end-to-end learning frameworks that directly model the mapping from hazy inputs to their corresponding clear counterparts.
For example, DEANet \cite{DEANet} improves dehazing performance through structural optimization and attention-based feature fusion, achieving a favorable trade-off between accuracy and computational efficiency.
DCMPNet \cite{DCMPNet} integrates depth estimation into the dehazing pipeline, leveraging depth cues to guide the restoration process, and demonstrates strong performance on synthetic datasets.
% While these efforts have led to significant advancements, they mostly focus on improving network structures to enhance performance on synthetically generated datasets, which ignores the diverse haze distribution in real-world images.
While these efforts have led to significant advancements, they primarily focus on enhancing performance on synthetic datasets, which often fail to perform well on complex real-world hazy images. In contrast, we incorporate patch-level analysis to improve generalization in handling complex and diverse haze distributions.

\section{Methods}
In this section, we first introduce AHG, which addresses the scarcity of training data to enhance generalization and mitigate domain gap. We then present DHR, which leverages patch importance learning to effectively handle complex and diverse haze distributions in real-world scenarios. Finally, we detail the loss functions employed to optimize the network.
\subsection{Automatic Haze Generation}

\begin{figure*}[t]
\centering
%\captionsetup{justification=justified, singlelinecheck=false}
\includegraphics[width=\linewidth]{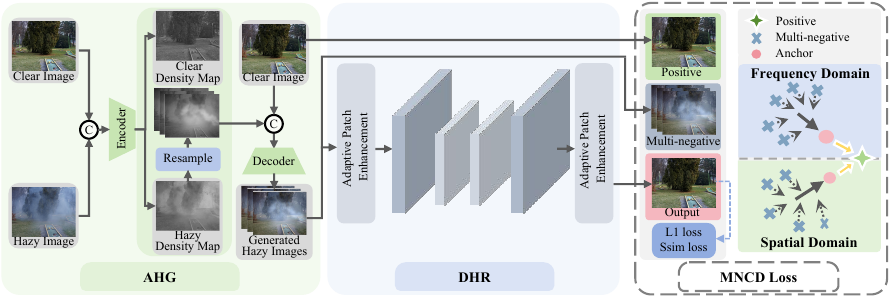} 
\caption{Overview of the proposed API framework. AHG adopts a hybrid data augmentation strategy to generate realistic and diverse hazy images, serving as the high-quality training data.
DHR further considers hazy regions with varying haze density distributions  in an adaptive patch importance-aware manner for generalizable real-world image dehazing.
}
\label{pipeline}
\end{figure*}
As illustrated in Figure~\ref{pipeline}, we propose AHG composed of an encoder $E$ and a decoder $D$. Given a pair of real hazy image $\mathbf{I}_\mathrm{h}$ and its corresponding clear image $\mathbf{I}_\mathrm{c}$, 
the encoder extracts single-channel haze density maps $\mathbf{M}_\mathrm{c}$ and $\mathbf{M}_\mathrm{h}$, and the decoder 
reconstructs both hazy and clear images from the clear input image and the corresponding density maps:
\begin{equation}
\mathbf{M}_\mathrm{c},\ \mathbf{M}_\mathrm{h} = E(\mathbf{I}_\mathrm{c},\ \mathbf{I}_\mathrm{h}),\quad
\widehat{\mathbf{I}}_x = D(\mathbf{I}_\mathrm{c},\ \mathbf{M}_x),\quad x \in \{\mathrm{c}, \mathrm{h}\},
\end{equation}
where $\widehat{\mathbf{I}}_x$ denotes the reconstructed image conditioned on the density map $\mathbf{M}_x$.

To supervise the reconstruction process, we apply an $\ell_1$ loss between the generated image $\widehat{\mathbf{I}}_x$ and the ground truth $\mathbf{I}_x$, and introduce an adversarial loss $\mathcal{L}_\mathrm{adv}$ to enhance the perceptual realism of the synthesized hazy image $\widehat{\mathbf{I}}_\mathrm{h}$. The overall training objective is:
\begin{equation}
\mathcal{L}_\mathrm{total} = \|\widehat{\mathbf{I}}_\mathrm{h} - \mathbf{I}_\mathrm{h}\|_1 + \lambda_1 \|\widehat{\mathbf{I}}_\mathrm{c} - \mathbf{I}_\mathrm{c}\|_1 + \lambda_2 \mathcal{L}_\mathrm{adv}(\widehat{\mathbf{I}}_\mathrm{h}, \mathbf{I}_\mathrm{h}).
\end{equation}
% where $\lambda_1 = 0.5$ and $\lambda_2 = 0.0001$ are empirically determined weights to balance the reconstruction and adversarial terms.

\paragraph{Hybrid Data Augmentation.}  
To enrich hazy training datasets, we resample the predicted density maps through a weighted interpolation between $\mathbf{M}_\mathrm{c}$ and $\mathbf{M}_\mathrm{h}$:
\begin{equation}
\widetilde{\mathbf{M}} = \alpha \mathbf{M}_\mathrm{c} + (1 - \alpha) \mathbf{M}_\mathrm{h}, \quad \alpha \in [0, 1],
\label{equ3}
\end{equation}
where $\alpha$ controls the haze intensity in the synthesized map $\widetilde{\mathbf{M}}$.
% Next, to impose spatially non-uniform haze distributions, we draw random offsets $u_i, v_i \sim \mathcal{U}(0, L)$
% for each octave and define normalized weights:
% \begin{equation}
%     w_i = \frac{p^i}{\sum_{j=0}^{O-1} p^j}\,.
% \end{equation}

% We then compute the multi-octave Perlin noise weight map as
% \begin{equation}
% \mathrm{PerlinDist}(x,y)
% = \sum_{i=0}^{O-1} w_i \,\mathrm{Perlin}\bigl(k^i x + u_i,\;k^i y + v_i\bigr)\,.
% \end{equation}

% Finally, the hybrid haze map is modulated by this distribution:
% \begin{equation}
% \widehat{\mathbf{M}}(x,y)
% = \widetilde{\mathbf{M}}(x,y)\,\odot\,\mathrm{PerlinDist}(x,y)\,,
% \label{equ8}
% \end{equation}
% and the decoder then uses \(\widehat{\mathbf{M}}\) to synthesize realistic hazy images with diverse spatial patterns and densities.
Next, to impose spatially non-uniform haze distributions, we resample the density map using Perlin noise \cite{p_noise}. Specifically, we generate multi-scale perturbations and compute a weighted sum across octaves as follows:
\begin{equation}
w_i = \frac{p^i}{\sum_{j=0}^{O-1} p^j},
\end{equation}
\begin{equation}
\mathrm{PerlinDist}(x,y) = \sum_{i=0}^{O-1} w_i\, \mathrm{Perlin}(k^i x + u_i,\; k^i y + v_i),
\end{equation}
where \( p \) is the persistence factor, \( k \) controls frequency scaling, and \( u_i, v_i \sim \mathcal{U}(0, L) \) are random offsets sampled from a uniform distribution. 

Finally, the hybrid haze density map $\widehat{\mathbf{M}}(x,y)$ is modulated by this spatial distribution:
\begin{equation}
\widehat{\mathbf{M}}(x,y)
= \widetilde{\mathbf{M}}(x,y)\, \odot\, \mathrm{PerlinDist}(x,y).
\label{equ5}
\end{equation}
The decoder then uses \(\widehat{\mathbf{M}}(x,y)\) to synthesize realistic hazy images with diverse haze distribution. More implementation details and visualizations of the simulated hazy images can be found in the supplementary material.
% We further introduce local perturbations to $\mathbf{M}_\mathrm{c}$, such as spatially varying Gaussian noise or region-specific weight adjustments, to simulate non-uniform haze.  
% The resulting perturbed maps are used by the decoder to generate realistic hazy images with diverse spatial distributions and haze densities.  
% This hybrid data augmentation strategy significantly enriches the training space, improving the robustness and generalization capability of the downstream dehazing network.

\subsection{Density-aware Haze Removal}

As illustrated in Figure~\ref{pipeline}, the proposed DHR is built upon a U-Net backbone.  
To enable adaptive and localized feature processing, we insert an Adaptive Patch Enhancement (APE) block both before and after passing through the U-Net backbone
% This design allows DHR to effectively model spatially varying haze patterns at the patch level while preserving global image consistency.

\begin{figure}
\label{APE}
\centering
\includegraphics[width=1.0\columnwidth]{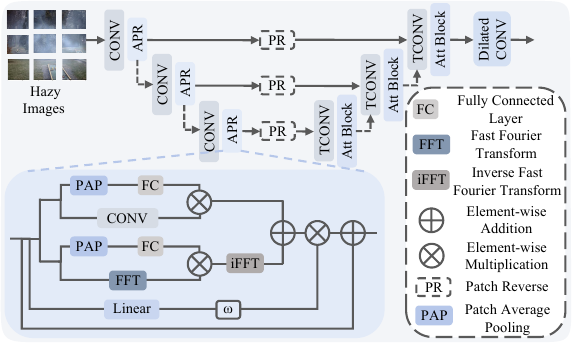}
\caption{Architecture of the Adaptive Patch Enhancement (APE). APE handles spatial and frequency patch-wise features separately, with adaptive residual connections applied to each patch.
}
\label{APE}
\end{figure}
% As shown in Figure~\ref{APE}, the APE block operates on an intermediate feature map $\mathbf{X} \in \mathbb{R}^{B \times C \times H \times W}$.  
% We begin by dividing $\mathbf{X}$ into a set of spatial patches, resulting in a reshaped tensor $\mathbf{P} \in \mathbb{R}^{N^2 B \times C \times \frac{H}{N} \times \frac{W}{N}}$, where $N^2$ denotes the number of patches per image.
% Each patch is then processed independently through a lightweight U-Net.  
% The encoder consists of a sequence of $3 \times 3$ convolutional layers with stride 2 for spatial reduction.
% At the bottleneck and during skip connections, a patch reversal operation is applied to restore the original spatial structure of the features followed by an Adapted Patch ResBlock (APR) to extract context-aware features specific to each patch.
% Subsequently, a transposed convolution-based decoder upsamples the feature maps, and attention modules are integrated to enhance informative regions. To further expand the receptive field and preserve fine details, we append a multi-scale dilated convolution block at the end of the APE. 

As shown in Figure~\ref{APE}, the APE block operates on an intermediate feature map $\mathbf{X} \in \mathbb{R}^{B \times C \times H \times W}$. We first divide $\mathbf{X}$ into a set of spatial patches, resulting in a reshaped tensor $\mathbf{P} \in \mathbb{R}^{N^2 B \times C \times \frac{H}{N} \times \frac{W}{N}}$, where $N^2$ denotes the number of patches per image. Each patch is then processed independently through a lightweight U-Net. The encoder consists of a series of $3 \times 3$ convolutional layers with stride 2 for spatial reduction. At the bottleneck and during skip connections, a patch reversal operation is applied to restore the original spatial structure of the features, followed by an Adapted Patch ResBlock (APR) to extract context-aware features for each patch. A transposed convolution-based decoder then upsamples the feature maps, and attention modules are integrated to emphasize informative regions. To further expand the receptive field and preserve fine details, we append a multi-scale dilated convolution block at the end of the APE.

\paragraph{Adapted Patch ResBlock}
As shown at the bottom left of Figure~\ref{APE}, the proposed APR adopts a dual-branch architecture that jointly models spatial and frequency information, followed by adaptive residual connections that integrate the outputs in a patch-aware manner.
% Both branches employ patch-wise channel attention. Unlike previous approaches such as \cite{FFA}, which compute channel attention over the entire image and may neglect local variations, our method performs attention computation at the patch level, enabling different patches to attend to distinct channel-wise features.

% \paragraph{Spatial Branch.} Given an input patch-level feature map $\mathbf{P}_\text{in}$, we first apply convolution, then compute channel-wise attention weights using patch average pooling (PAP) followed by a fully connected layer (FC).  
% The resulting weights are multiplied with the feature map along the channel dimension:
% \begin{equation}
% \mathbf{P}_\text{spa} = \text{Conv}(\mathbf{P}_\text{in}) \otimes \text{FC} \big( \text{PAP}(\mathbf{P}_\text{in}) \big),
% \end{equation}
% where $\otimes$ denotes channel-wise multiplication.

% \paragraph{Frequency Branch.} In parallel, the frequency branch applies a Fast Fourier Transform (FFT) \cite{fft} to project the input into the frequency domain.  
% After attention weighting (using the same PAP-FC structure), the result is transformed back to the spatial domain via an inverse FFT (iFFT):
% \begin{equation}
% \mathbf{P}_\text{fre} = \text{iFFT} \Big( \text{FFT}(\mathbf{P}_\text{in}) \otimes \text{FC} \big( \text{PAP}(\mathbf{P}_\text{in}) \big) \Big),
% \end{equation}

% \paragraph{Spatial and Frequency Branches.}
Given an input patch-level feature map $\mathbf{P}_\text{in}$, the spatial branch applies convolution followed by channel-wise attention via patch average pooling (PAP) and a fully connected layer (FC), while the frequency branch transforms the input into the frequency domain via FFT, applies the same PAP-FC attention, and then reconstructs the spatial representation using inverse FFT. The two branches are formulated as:
\begin{equation}
\mathbf{P}_\text{spa} = \text{Conv}(\mathbf{P}_\text{in}) \otimes \text{FC} \big( \text{PAP}(\mathbf{P}_\text{in}) \big),
\end{equation}
\begin{equation}
\mathbf{P}_\text{fre} = \text{iFFT} \Big( \text{FFT}(\mathbf{P}_\text{in}) \otimes \text{FC} \big( \text{PAP}(\mathbf{P}_\text{in}) \big) \Big),
\end{equation}
where $\otimes$ denotes channel-wise multiplication.

% \paragraph{Adaptive Residual Fusion.} 
% While residual connections~\cite{MSBDN} are widely used in dehazing networks, they often treat all spatial regions equally.  
% However, in real-world scenarios, different patches may require different levels of processing—heavily hazy regions need stronger refinement, while relatively clear regions may only require identity mapping.

% To address this, we design an adaptive residual fusion mechanism.  
% We compute a scalar weight $w$ for each patch via a linear transformation followed by a sigmoid function applied to the pooled sum of the two branches:
% \begin{equation}
% w = \sigma \Big( \mathbf{W} \cdot \text{PAP}(\mathbf{P}_\text{spa} + \mathbf{P}_\text{fre}) \Big),
% \end{equation}

% where $W$ is the weight matrix of a linear layer and $\sigma(\cdot)$ denotes the sigmoid function.

% Finally, the output of the APR block is computed by fusing the original input and the processed features from both branches in a weighted residual manner:
% \begin{equation}
% \mathbf{P}_\text{out} = w \cdot \mathbf{P}_\text{in} + (1 - w) \cdot \big( \mathbf{P}_\text{spa} + \mathbf{P}_\text{fre} \big).
% \end{equation}

To address spatially non-uniform haze in real world, we design an adaptive residual fusion mechanism. A scalar weight $w$ is computed via a linear transformation and sigmoid activation on the pooled sum of the two branches, and the final output is fused in a weighted residual manner:
\begin{equation}
w = \sigma \Big( \mathbf{W} \cdot \text{PAP}(\mathbf{P}_\text{spa} + \mathbf{P}_\text{fre}) \Big),
\end{equation}
\begin{equation}
\mathbf{P}_\text{out} = w \cdot \mathbf{P}_\text{in} + (1 - w) \cdot \big( \mathbf{P}_\text{spa} + \mathbf{P}_\text{fre} \big),
\end{equation}
where $\mathbf{W}$ is the weight matrix of a linear layer and $\sigma(\cdot)$ denotes the sigmoid function.

\begin{table*}[t]
\fontsize{7pt}{9.2pt}\selectfont
\caption{
Quantitative comparison between our method and state-of-the-art approaches on four real-world paired datasets using SSIM and PSNR metrics.
The best result is highlighted in \textbf{bold}, and the second best is \underline{underlined}.
}
\label{table_paired}
\centering
\resizebox{1.0\linewidth}{!}{
\begin{tabular}{c c|c c|c c|c c|c c}
\hline
\multicolumn{2}{c|}{\multirow{2}{*}{Method}} &\multicolumn{2}{c|}{Dense-Haze \cite{DENSE-HAZE}}&\multicolumn{2}{c|}{NH-Haze \cite{NH-HAZE}}& \multicolumn{2}{c|}{O-Haze \cite{O-HAZE}}&\multicolumn{2}{c}{I-Haze \cite{I-HAZE}} \\
\cline{3-10}
\multicolumn{2}{c|}{} & PSNR$\uparrow$ & SSIM$\uparrow$ & PSNR$\uparrow$ & SSIM$\uparrow$ & PSNR$\uparrow$ & SSIM$\uparrow$ & PSNR$\uparrow$ & SSIM$\uparrow$  \\
\hline

\multicolumn{2}{c|}{GridDehazeNet \cite{griddehazenet}} & 13.447 &0.665 &18.278 &0.745 &22.445 &0.910 &19.078&0.883 \\ 

\multicolumn{2}{c|}{FFA \cite{FFA} } & 14.086 &0.686 &19.788 &0.763 &23.321 &0.924 &20.566 &0.915 \\ 

\multicolumn{2}{c|}{MSBDN \cite{MSBDN}} & 14.123 &0.695 &19.400 &0.772 &23.888 &0.927 &20.960&0.916  \\ 

\multicolumn{2}{c|}{DeHamer \cite{Dehamer}} &15.278 &0.700 &19.520 &0.743 & 24.505 & 0.926 & 21.346 &0.917 \\ 

\multicolumn{2}{c|}{C2P\cite{c2p}} &15.355 &0.695 &19.437 &0.782 & 24.068 & 0.942 & 19.947 &0.911\\ 

\multicolumn{2}{c|}{Fourmer\cite{fourmer} } & 15.428 &0.702 &19.345 &0.763 &23.831 &0.942 &20.044 &0.909 \\ 

\multicolumn{2}{c|}{DehazeFormer \cite{dehazeformer} } & 15.484&\underline{0.706} &20.236 &0.775 &24.227 &0.936 &21.398 &\underline{0.923}  \\ 

\multicolumn{2}{c|}{MITNet \cite{MITNet} } &15.235 &0.695 &20.129 &0.756 &24.157 &\underline{0.947} & 20.365& 0.918   \\ 

\multicolumn{2}{c|}{DEANet \cite{DEANet} } &\underline{15.868} & 0.687&\underline{20.291} &\underline{0.784} &\underline{24.618}  &0.945 &21.681 & 0.918 \\ 

\multicolumn{2}{c|}{DCMPNet \cite{DCMPNet} } & 15.120&0.671 &19.891 &0.763 &24.457  &0.935  &\underline{21.956} &0.915   \\ 

\hline
\multicolumn{2}{c|}{RIDCP \cite{RIDCP}} &9.587 &0.437 &13.472 &0.461 &17.540 &0.649  &16.066 &0.718  \\

\multicolumn{2}{c|}{CORUN \cite{CORUN}} &9.893 &0.494 &12.184 &0.576 &17.865 &0.684  &17.305 &0.764  \\ 

\multicolumn{2}{c|}{PTTD \cite{PTTD}} &14.537 &0.498 &15.595 &0.571 &19.473 &0.674  &17.046 &0.746  \\ 

\multicolumn{2}{c|}{IPC \cite{IPC}} &7.138 &0.414 &12.101 &0.541 &16.908 &0.7031  &16.486 &0.739  \\ 

\hline

\multicolumn{2}{c|}{Ours} 
&\textbf{16.184}
&\textbf{0.720} &\textbf{20.652} &\textbf{0.803} &\textbf{25.960} &\textbf{0.954} &\textbf{23.158} &\textbf{0.934}  \\ 
\hline
\end{tabular}}
\end{table*}

This design enables patch-wise modulation of residual strength, thereby allowing the network to adaptively refine features based on local haze severity.  
More implementation details of DHR, along with visualizations of the weights assigned to different patches, can be found in the supplementary material.

\subsection{Training Losses}
Recent advances in contrastive learning \cite{simclr} have been successfully applied to low-level vision tasks~\cite{contrast}. However, most existing methods associate each anchor with only a single negative sample, limiting the diversity of contrastive information. In contrast, we leverage AHG to generate diverse hazy images, allowing the construction of multiple negative samples for each anchor. 
Additionally, we extend the contrastive loss computation to both the spatial and frequency domains.
Specifically, we use a pre-trained VGG-16 \cite{vgg} network as the spatial-domain feature extractor $E(\cdot)$, and Fast Fourier Transform $F(\cdot)$ to capture frequency-domain features.  
For an anchor image $\mathbf{A}$, a positive sample $\mathbf{P}$, and a set of negative samples $\{\mathbf{N}_i\}_{i=1}^{n}$, the per-pair contrastive loss is defined as:
\begin{equation}
\mathcal{L}(\mathbf{A}, \mathbf{X}) = \|\!| E(\mathbf{A}) - E(\mathbf{X}) \|\!|_1 + \|\!| F(\mathbf{A}) - F(\mathbf{X}) \|\!|_1,
\end{equation}
where $\mathbf{X} \in \{\mathbf{P}, \mathbf{N}_i\}$ and $\|\!|\cdot\|\!|_1$ denotes the $\ell_1$ distance.
The final Multi-Negative Contrastive Dehazing (MNCD) loss is formulated as:
\begin{equation}
\mathcal{L}_\text{MNCD} = \frac{\mathcal{L}(\mathbf{A}, \mathbf{P})}{\mathcal{L}(\mathbf{A}, \mathbf{P}) + \sum_{i=1}^{n} \lambda_i \mathcal{L}(\mathbf{A}, \mathbf{N}_i)},
\end{equation}
where $\lambda_i$ denotes the importance weight of the $i$-th negative sample.  
We set $\lambda_i = 1$ for real hazy negatives and $\lambda_i = 0.5$ for generated negatives.  
In practice, we sample $n=10$ negative examples for each training instance.

% \paragraph{Overall Training Loss.}  
The final joint loss combines the MNCD loss with a smooth $\ell_1$ reconstruction loss and the multi-scale structural similarity (MS-SSIM) loss.  
The overall loss function is defined as:
\begin{equation}
\mathcal{L}_\text{joint} = \lambda_1 \mathcal{L}_{\text{SL1}} + \lambda_2 \mathcal{L}_{\text{MS-SSIM}} + \lambda_3 \mathcal{L}_{\text{MNCD}},
\end{equation}
where $\lambda_1$, $\lambda_2$, and $\lambda_3$ are empirically set to 1, 0.5, and 0.05, respectively.

% \begin{figure*}[t]
% \centering
% \includegraphics[width=0.99\columnwidth]{figures/paired.pdf} 
% \caption{A qualitative comparison between our method and state-of-the-art methods on four paired real-world datasets.}
% \label{paired_compare}
% \end{figure*}
\begin{figure*}[t]
\centering
\includegraphics[width=2.1\columnwidth]{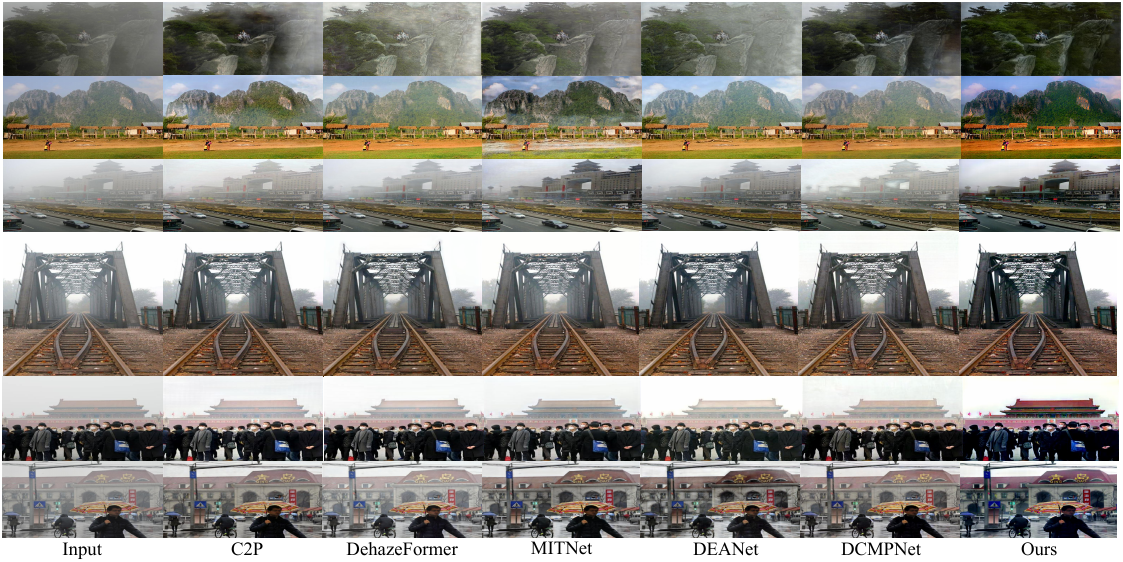} 
\caption{A qualitative comparison between our method and state-of-the-art methods on real-world unpaired dataset Fattal \cite{fattal}, RTTS  \cite{SOTS} and URHI \cite{SOTS}.}
\label{qualitative_comparison}
\end{figure*}

\section{Experiments}
\subsection{Experimental Setting}
\paragraph{Implementation Details.}  
All experiments are conducted on an NVIDIA GeForce RTX 4090 GPU.  
The model is implemented using the PyTorch framework and optimized with the Adam optimizer.  
We adopt a cosine annealing learning rate scheduler with an initial learning rate of $1 \times 10^{-4}$ and a minimum learning rate of $1 \times 10^{-6}$.  
The batch size is set to 4.  
Training images are randomly cropped to sizes that are multiples of 256 and then resized to $256 \times 256$.  
The network is trained for a total of 400 epochs.
\paragraph{Datasets.}  
We train our model on four real-world hazy datasets: NH-Haze~\cite{NH-HAZE}, Dense-Haze~\cite{DENSE-HAZE}, I-Haze~\cite{I-HAZE}, and O-Haze~\cite{O-HAZE}, as well as one synthetic datasets SOTS ~\cite{SOTS}.  
During training, input samples are randomly selected from the original datasets or generated using the proposed AHG module.  
The sampling probability of AHG-generated images gradually decreases as training progresses, encouraging the model to focus more on real data in later stages.  
We evaluate model performance on the four aforementioned real-world paired datasets and three additional real-world unpaired dataset, Fattal~\cite{fattal}, RTTS \cite{SOTS} and URHI \cite{SOTS}, to assess generalization to unseen real-world haze distributions.
\paragraph{Compared Methods.}  
We compare our approach against a broad range of state-of-the-art dehazing methods, evaluating them in two settings: (1) training them on our datasets using their official implementations, including GridDehazeNet~\cite{griddehazenet}, FFA-Net~\cite{FFA}, MSBDN~\cite{MSBDN}, DeHamer~\cite{Dehamer}, C2P~\cite{c2p}, Fourmer~\cite{fourmer}, DehazeFormer~\cite{dehazeformer}, MITNet~\cite{MITNet}, 
DEANet~\cite{DEANet}, and DCMPNet~\cite{DCMPNet}, and (2) using their officially released pretrained weights without further finetuning, including RIDCP~\cite{RIDCP}, CORUN~\cite{CORUN}, PTTD~\cite{PTTD}, and IPC~\cite{IPC}.

\begin{figure*}[t]
\centering
%\captionsetup{justification=justified, singlelinecheck=false}
\includegraphics[width=\linewidth]{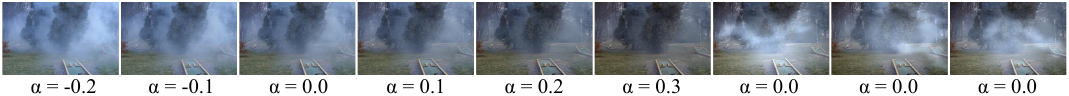} 
\caption{Visualization of hazy images generated by AHG. The first six examples illustrate the effect of varying haze intensity controlled by Equation~\ref{equ3}, while the last three demonstrate the modulation of spatial haze distribution via Equation~\ref{equ5}.}
\label{vis_haze}
\end{figure*}

% As shown in Table~\ref{table_paired}, we quantitatively evaluate the dehazing performance of our model on four real-world paired hazy datasets using two standard metrics: Structural Similarity Index Measure (SSIM) and Peak Signal-to-Noise Ratio (PSNR) \cite{psnr}.  
% Our method consistently achieves superior performance across all datasets when compared to state-of-the-art approaches.
% Specifically, our model surpasses the best existing methods by 0.014, 0.019, 0.007, and 0.011 in SSIM on the Dense-Haze~\cite{DENSE-HAZE}, NH-Haze~\cite{NH-HAZE}, O-Haze~\cite{O-HAZE}, and I-Haze~\cite{I-HAZE}datasets, respectively.  
% In terms of PSNR, our approach outperforms prior methods by 0.316~dB, 0.361~dB, 1.342~dB, and 1.202~dB on the same datasets.
% These results clearly demonstrate the effectiveness and generalizability of our method across diverse real-world scenarios.

\subsection{Quantitative Evaluation}
As shown in Table~\ref{table_paired}, we quantitatively evaluate the dehazing performance of our proposed model across four real-world paired hazy datasets using two widely adopted metrics: Structural Similarity Index Measure (SSIM) \cite{ssim} and Peak Signal-to-Noise Ratio (PSNR) \cite{psnr}.

Our method consistently demonstrates superior performance compared to state-of-the-art approaches across all evaluated datasets. Specifically, our model achieves notable improvements in SSIM scores, surpassing the best previous methods by 0.014, 0.019, 0.007, and 0.011 on Dense-Haze~\cite{DENSE-HAZE}, NH-Haze~\cite{NH-HAZE}, O-Haze~\cite{O-HAZE}, and I-Haze~\cite{I-HAZE}, respectively. These improvements highlight our model's ability to preserve structural fidelity and perceptual quality of the restored images.
In terms of PSNR, our method delivers even more significant gains, outperforming prior methods by 0.316dB, 0.361dB, 1.342dB, and 1.202dB on the corresponding datasets. Particularly noteworthy is the substantial improvement observed on the O-Haze and I-Haze datasets, where our approach significantly exceeds existing state-of-the-art methods by more than 1~dB. Such quantitative advancements confirm our method’s superior ability in reconstructing visually accurate and detailed images from hazy conditions.

\begin{table}
\centering
\caption{Ablation study on our proposed APE, AHG, and MNCD loss.} 
\label{tab:ablation_module}
\resizebox{1.0\linewidth}{!}{
\begin{tabular}{c|c c c c|c c}
\hline
Models & APE  & AHG & MNCD &  &PSNR$\uparrow$ & SSIM$\uparrow$ \\
\hline
Baseline &  &  &  &  &19.322  &0.814   \\
DHR & \checkmark &  &  &  &20.254  &0.834   \\
DHR+AHG & \checkmark   & \checkmark &  & & 21.019& 0.842 \\
API & \checkmark   & \checkmark &\checkmark  & & \textbf{21.488} & \textbf{0.852} \\
% & \checkmark   & \checkmark &\checkmark  &\checkmark &22.72 & 0.832 \\
\hline
\end{tabular}}
\end{table}

\subsection{Qualitative Evaluation}
We further evaluate our model qualitatively on real-world unpaired datasets to verify its practical effectiveness. As illustrated in Figure~\ref{qualitative_comparison}, our proposed method consistently achieves superior haze removal results compared to existing approaches. Specifically, the images recovered by our model exhibit significantly clearer structures, improved visibility of fine details, higher overall contrast, and more accurate, natural color tones.
Moreover, our method demonstrates remarkable visual consistency, effectively handling regions with varying haze densities—from densely hazy areas to relatively clear regions. These qualitative improvements underscore our model's excellent generalization capabilities when applied to complex real-world scenarios characterized by diverse haze distributions.

As shown in Figure \ref{vis_haze}, we visualize hazy images generated by our AHG. The results demonstrate that both haze intensity and spatial distribution are well controlled, leading to highly realistic and visually convincing synthetic hazy images.

Overall, these results robustly demonstrate the effectiveness and generalization capability of our approach, underscoring its practical applicability across diverse real-world dehazing scenarios.
Additional qualitative comparisons based on real-world paired datasets and synthetic datasets SOTS \cite{SOTS} are provided in the supplementary material for further reference.

\subsection{Ablation Study}

\paragraph{Effectiveness of Architectures.}
To evaluate the effectiveness of each component in the proposed API framework, we conduct ablation studies on three key elements: APE module, AHG module, and MNCD loss. As shown in Table~\ref{tab:ablation_module}, we progressively integrate these components into the baseline and report the average PSNR and SSIM across four real-world paired datasets. Each module yields consistent performance gains, validating its individual contribution. Together, they form a complementary and synergistic framework that significantly enhances overall dehazing performance.

% To further evaluate APE, we provide a qualitative comparison of dehazing results with and without APE.  
% As shown in Figure~\ref{ablation_APE}, incorporating APE results in cleaner and more visually pleasing reconstructions, particularly in scenes with dense or spatially complex haze.  
% These results highlight APE’s ability to enhance local feature representation and improve robustness under challenging haze distributions.

\begin{table}
\centering
\caption{Ablation study on the $\lambda_3$ in MNCD Loss}
\label{tab:ablation_MNCD}
\resizebox{1.0\linewidth}{!}{
\begin{tabular}{c|ccccc}
\hline
$\lambda_3$ & 1 & 0.5 & 0.05 & 0.01 & 0.005 \\
\hline
AVG PSNR$\uparrow$ & 20.959 & \underline{21.453} & \textbf{21.488} & 21.015 & 20.899 \\
AVG SSIM$\uparrow$ & 0.841 & 0.839 & \textbf{0.849} & \underline{0.842} & 0.840 \\
\hline
\end{tabular}}
\end{table}

\paragraph{Effectiveness of MNCD Loss.}
We also investigate the effect of the weight $\lambda_3$ in the MNCD Loss.  
As shown in Table~\ref{tab:ablation_MNCD}, our selected setting achieves the best performance, indicating the importance of properly weighting the loss components to guide the model effectively.

\paragraph{Impact of Patch Size.}
To examine the role of patch-based processing in APE, we perform an ablation study on different patch sizes.  
As reported in Table~\ref{tab:ablation_SIZE}, the best performance is achieved with a patch size of 32, which balances local detail modeling and global consistency.  
Larger patch sizes reduce local adaptivity and lead to performance degradation, while excessively small patches limit the receptive field, also resulting in suboptimal results.  

\begin{table}
\centering
\caption{Ablation study on the patch size.
} 
\label{tab:ablation_SIZE}
\resizebox{1.0\linewidth}{!}{
\begin{tabular}{c|c c c c c}
\hline
PATCH{\_}SIZE & 8 & 16 & 32 & 64 & 128 \\
\hline
AVG PSNR$\uparrow$ &20.854  &\underline{20.922} & \textbf{21.488}&20.121&19.712 \\
AVG SSIM$\uparrow$ &\underline{0.847}  &0.841 &\textbf{0.852}&0.831&0.821\\
% & \checkmark   & \checkmark &\checkmark  &\checkmark &22.72 & 0.832 \\
\hline
\end{tabular}}
\end{table}

\section{Conclusion}
In this paper, we propose \textit{API}, an \textbf{A}daptive \textbf{P}atch \textbf{I}mportance-aware paradigm for robust real-world image dehazing.  
Our framework begins with the Automatic Haze Generator (AHG), which expands the diversity of training data through controllable haze simulation.  
We then introduce the Density-aware Haze Removal (DHR) module, which performs dynamic, patch-based dehazing to effectively handle spatially varying haze.  
In addition, we present a Multi-Negative Contrastive Dehazing (MNCD) loss that leverages multiple negative samples to enhance feature discrimination in both the spatial and frequency domains.
Together, these components enable a single model to robustly address a wide range of real-world dehazing scenarios, significantly outperforming previous state-of-the-art methods on multiple benchmarks.  
We believe that our work offers a valuable contribution to the field of real-world image dehazing and hope it inspires further research into generalizable image restoration in complex environments.

{
    \small
    \bibliographystyle{ieeenat_fullname}
    \bibliography{main}
}

\clearpage

\appendix
In this supplementary material, we provide additional details and results to support the main paper.
Section A presents further implementation details of our model, including the design of the AHG and DHR modules.
Section B provides extended qualitative results: Figure \ref{paired_compare} shows visual comparisons on real-world paired datasets; Figure \ref{sot} presents results on synthetic datasets; Figure \ref{retults11} visualizes the patch-wise attention weights assigned by the APR; and Figures \ref{AHG1} and \ref{AHG2} illustrate the hazy images generated by the AHG module.

\section{Model Details}

\subsection{AHG Details}

In the main paper, we introduced the Automatic Haze Generation (AHG) module and its role in generating large quantities of simulated hazy images to support the training of the dehazing network.  
Here, we provide a detailed description of the network architecture and training strategy used in AHG.

The AHG module is implemented as a compact U-Net architecture, where both the encoder and decoder are composed of a series of residual blocks.  
Each residual block consists of a Conv-ReLU-Conv structure, and all downsampling and upsampling operations are performed using $3 \times 3$ convolutions or transposed convolutions with a stride of 2.  
Following each of these operations, we apply two residual blocks to refine the features.  
In total, the encoder performs two downsampling operations, and the decoder performs two corresponding upsampling operations, resulting in a bottleneck feature map that is one-quarter the resolution of the input.

To extract haze-related features, the input to the encoder is formed by concatenating a real hazy image and its corresponding clear image.  
After encoding, the network outputs two single-channel haze density maps using two $3 \times 3$ convolutional layers—one corresponding to the hazy input and one to the clear input.  
These maps represent the estimated haze concentration distribution in each image.

The decoder takes as input the concatenation of a density map and a clear image.  
The density map encodes the haze distribution, while the clear image provides structural and textural information needed for haze synthesis.  
During training, the decoder is supervised to reconstruct both hazy and clear images from their corresponding density maps.  
During inference, we modify the density map through resampling to produce a wide variety of haze distributions.  
This allows AHG to generate diverse, realistic hazy images with controllable haze intensity and spatial variation, significantly enhancing the diversity of the training dataset.

\subsection{DHR Architecture}

In the main text, we introduced the key components of the Density-aware Haze Removal (DHR) module, including the Adaptive Patch Enhancement (APE) and Adapted Patch ResBlock (APR).  
Here, we provide a detailed description of the full architecture of DHR and its supporting components.

DHR is built upon a U-Net-style encoder-decoder structure.  
We use $3 \times 3$ convolutions with a stride of 2 for downsampling and $4 \times 4$ transposed convolutions with a stride of 2 for upsampling.  
Each downsampling and upsampling operation is followed by a dilated block to capture multi-scale contextual information.  
The first downsampling layer and the last upsampling layer are equipped with APE blocks to provide adaptive local enhancement, while standard attention blocks are applied to the intermediate layers.  
The architecture consists of three downsampling and three upsampling stages, with skip connections linking corresponding encoder and decoder layers.

The total number of parameters in the DHR module is approximately 14M, with a computational cost of around 20 GMACs when processing $128 \times 128$ images.  
Compared to other methods, DHR achieves a favorable trade-off between modeling capacity and computational efficiency, allowing it to robustly handle a wide variety of real-world haze conditions with minimal overhead.

\paragraph{Attention Block.}
Attention blocks are used throughout both the APE and backbone components of DHR to enhance feature representation.  
Each attention block begins with a $3 \times 3$ convolution followed by an activation function.  
We then apply channel attention to emphasize informative feature channels, followed by pixel-level attention to refine local details.  
The resulting attention-modulated features are combined with the original processed feature map through a residual connection to form the final output.

\paragraph{Dilated Block.}
To increase the receptive field and capture features at multiple scales, each dilated block applies four parallel dilated convolutions with dilation rates of 1, 2, 3, and 4, respectively.  
The output channel allocation for each branch is as follows: $n - 3 \times \lfloor n / 10 \rfloor$ for the first branch, and $\lfloor n / 10 \rfloor$ for each of the remaining three branches, where $n$ is the number of input channels.  
These four outputs are concatenated along the channel dimension and then passed through an attention block for feature fusion.

We also plan to design dynamic network architectures \cite{ea-vit} that can adaptively adjust to low-level image restoration tasks, enabling more efficient and flexible image restoration methods. This direction will be explored in future work.

\section{More Results}

\begin{figure*}[t]
\centering
\includegraphics[width=2\columnwidth]{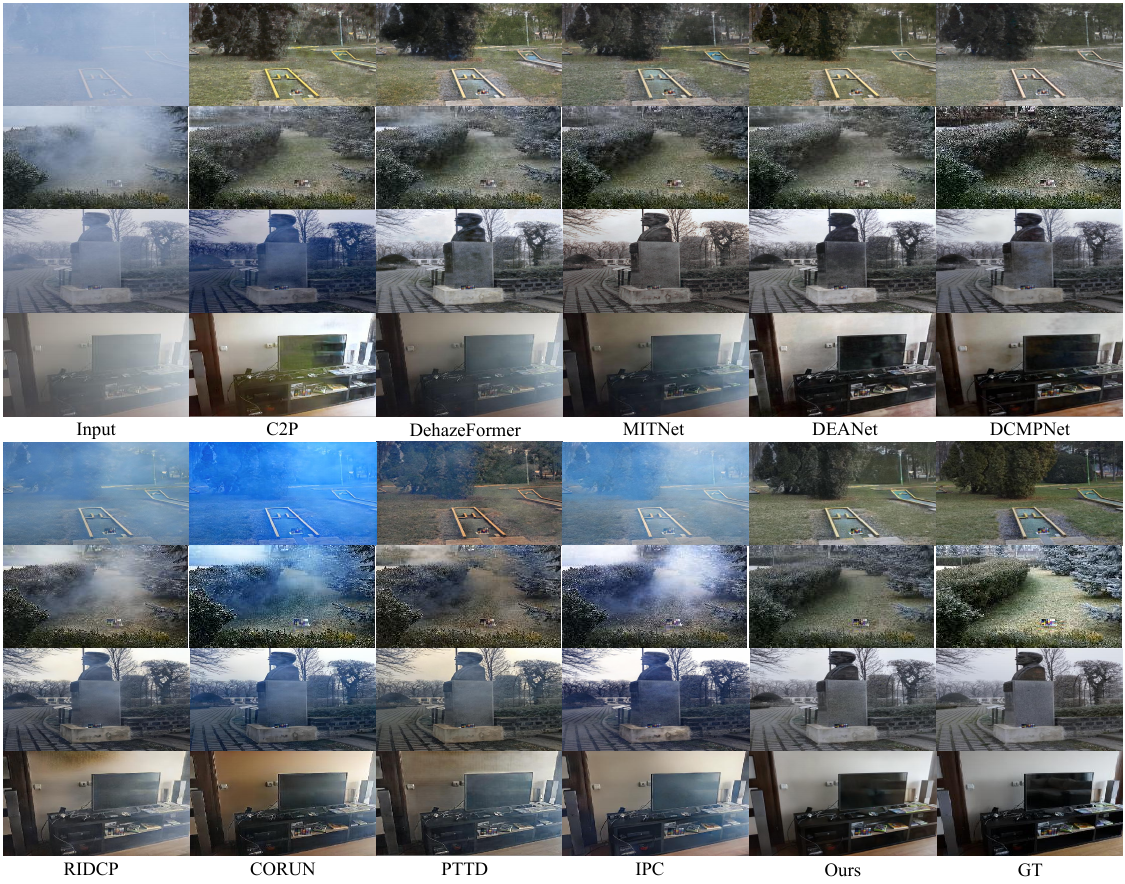} 
\caption{A qualitative comparison between our method and state-of-the-art methods on four paired real-world datasets.}
\label{paired_compare}
\end{figure*}

\paragraph{Qualitative Evaluation on Paired Dataset.}  
To further verify our method's effectiveness, we perform qualitative evaluations on paired real-world datasets. As shown in Figure~\ref{paired_compare}, our approach produces significantly clearer and more visually appealing images compared to state-of-the-art methods. Specifically, our method better preserves structural details, achieves superior color restoration, and generates more natural transitions across regions with varying haze densities. These visual enhancements demonstrate that our framework not only quantitatively surpasses existing methods but also consistently yields higher perceptual quality, highlighting its practical value for real-world image dehazing.

\begin{figure*}[t]
\centering
\includegraphics[width=1.99\columnwidth]{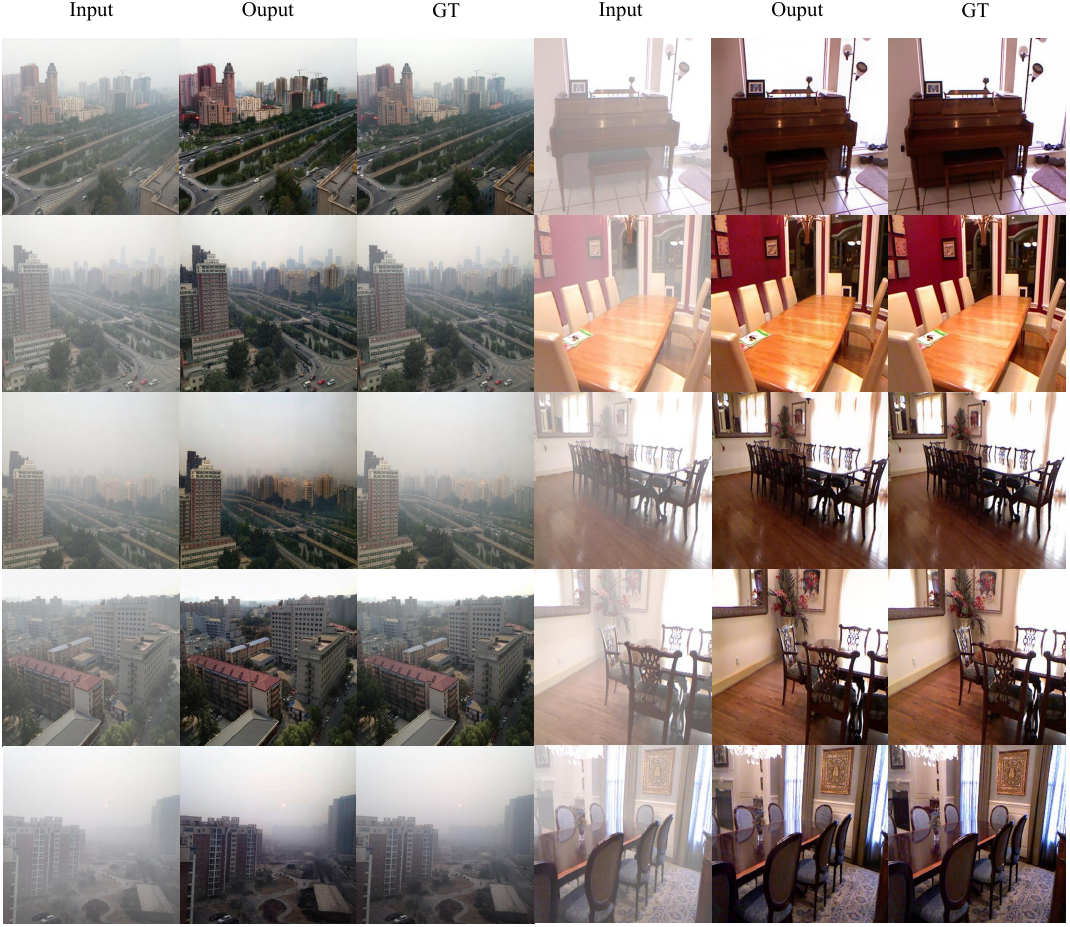} 
\caption{
Dehazing performance of our model on the SOTS-Indoor(right) and SOTS-Outdoor(left) datasets. Our method also achieves superior dehazing results on these artificially simulated hazy images using atmospheric scattering model, particularly on the SOTS-Outdoor dataset, where it even surpasses the dehazing of original clear images. Similar results are observed on the SOTS-Indoor test set.
}
\label{sot}
\end{figure*}

\paragraph{Results on Synthetic Datasets.}  
To evaluate the generalization ability of our model, we also test it on the commonly used synthetic dehazing benchmarks: SOTS-Indoor and SOTS-Outdoor.  
Both datasets simulate hazy images using the atmospheric scattering model.  
As shown in Figure~\ref{sot}, our model achieves highly effective dehazing on SOTS-Outdoor, in some cases even producing results that are visually cleaner than the provided "clear" ground truth images.  
This may be due to the fact that the ground truth images were captured under naturally hazy conditions, and artificial haze was subsequently added.  
These observations highlight a critical limitation of prior works that overfit to synthetic datasets: although they perform well on simulated benchmarks, they may struggle in real-world scenarios.  
In contrast, our method maintains robust performance across both synthetic and real-world domains.

Similarly, on the SOTS-Indoor test set, our model also achieves impressive results, demonstrating its strong adaptability to varying haze conditions in controlled environments.

\begin{figure*}[t]
  \centering
  \begin{subfigure}[t]{0.48\textwidth}
    \centering
    \includegraphics[width=\linewidth]{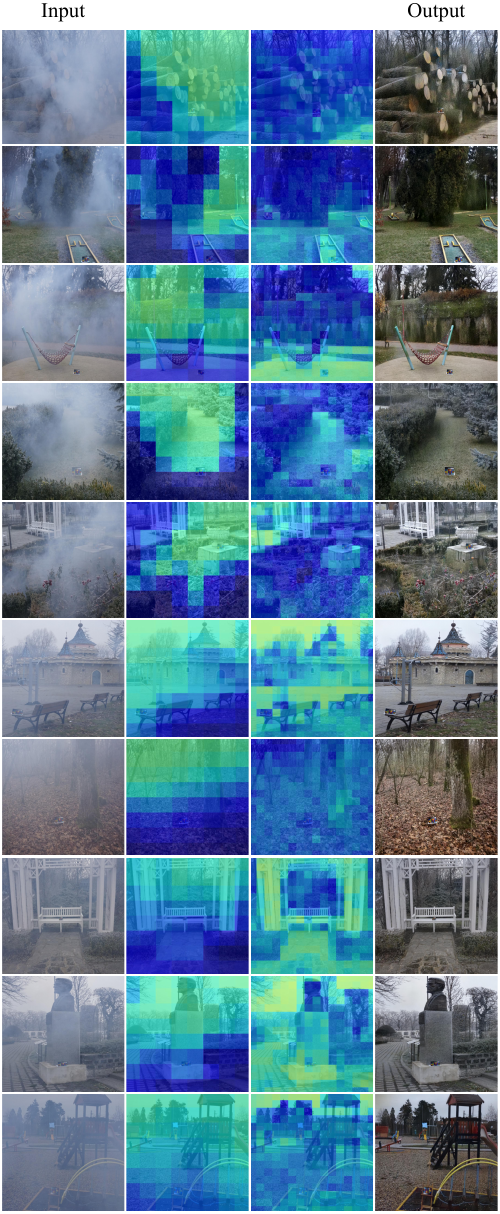}
    \caption{results 1}
    \label{retults2}
  \end{subfigure}
  \hfill
  \begin{subfigure}[t]{0.48\textwidth}
    \centering
    \includegraphics[width=\linewidth]{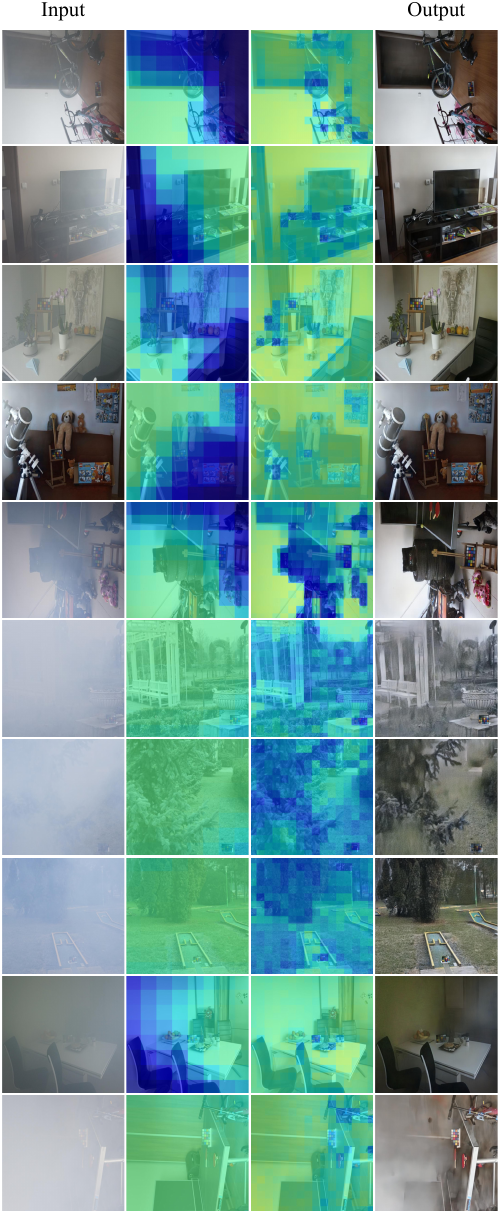}
    \caption{results 2}
    \label{retults1}
  \end{subfigure}
  \caption{Additional dehazing result on four paired datasets, accompanied by the corresponding patch weight distributions in the Adaptive Patch ResBlock (APR). As shown, our model achieves impressive dehazing performance, and APR effectively differentiates between patches, handling varying haze concentrations across the images.}
  \label{retults11}
\end{figure*}

\paragraph{Visualizations of Patch-wise Attention Weights from APR.}
In addition to the examples presented in the main paper, we provide further dehazing results on four real-world paired datasets, along with corresponding visualizations of patch-wise attention weights from the Adapted Patch ResBlock (APR).
As illustrated in Figure~\ref{retults11}, our model consistently produces high-quality dehazed images, while APR effectively distinguishes patches with varying haze intensities.
These visualizations highlight the model’s capacity for adaptive local enhancement and demonstrate its ability to handle spatially heterogeneous haze distributions commonly encountered in real-world scenarios.

\begin{figure*}[t]
\centering
\includegraphics[width=1.99\columnwidth]{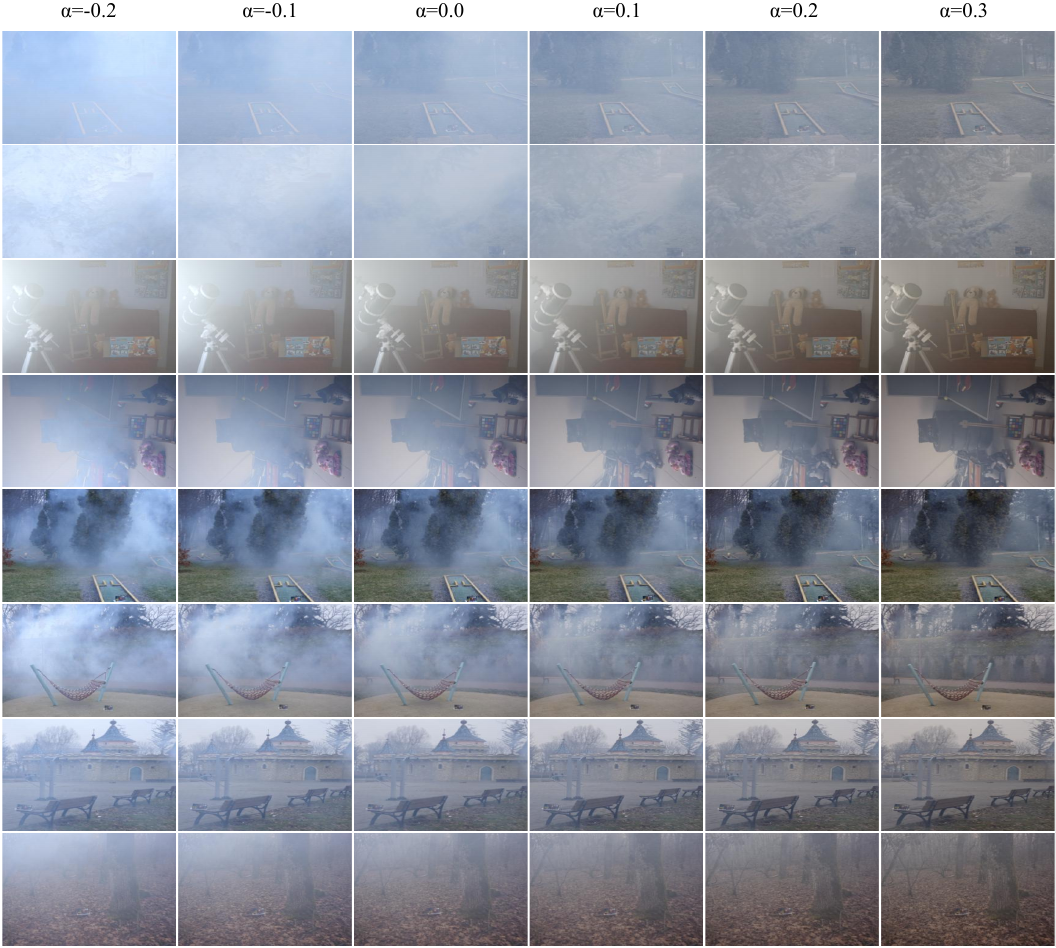} 
\caption{
Demonstration of hazy image generation by varying the coefficient $\alpha$ from -0.2 to 0.3 using Equal 3. The figure shows the resulting hazy images for four paired datasets, with two sets selected for each dataset. By adjusting $\alpha$, we generate different haze distributions, ranging from dense haze to light haze, allowing us to simulate various haze levels and types in the generated images.
}
\label{AHG1}
\end{figure*}

\begin{figure*}[t]
\centering
\includegraphics[width=1.99\columnwidth]{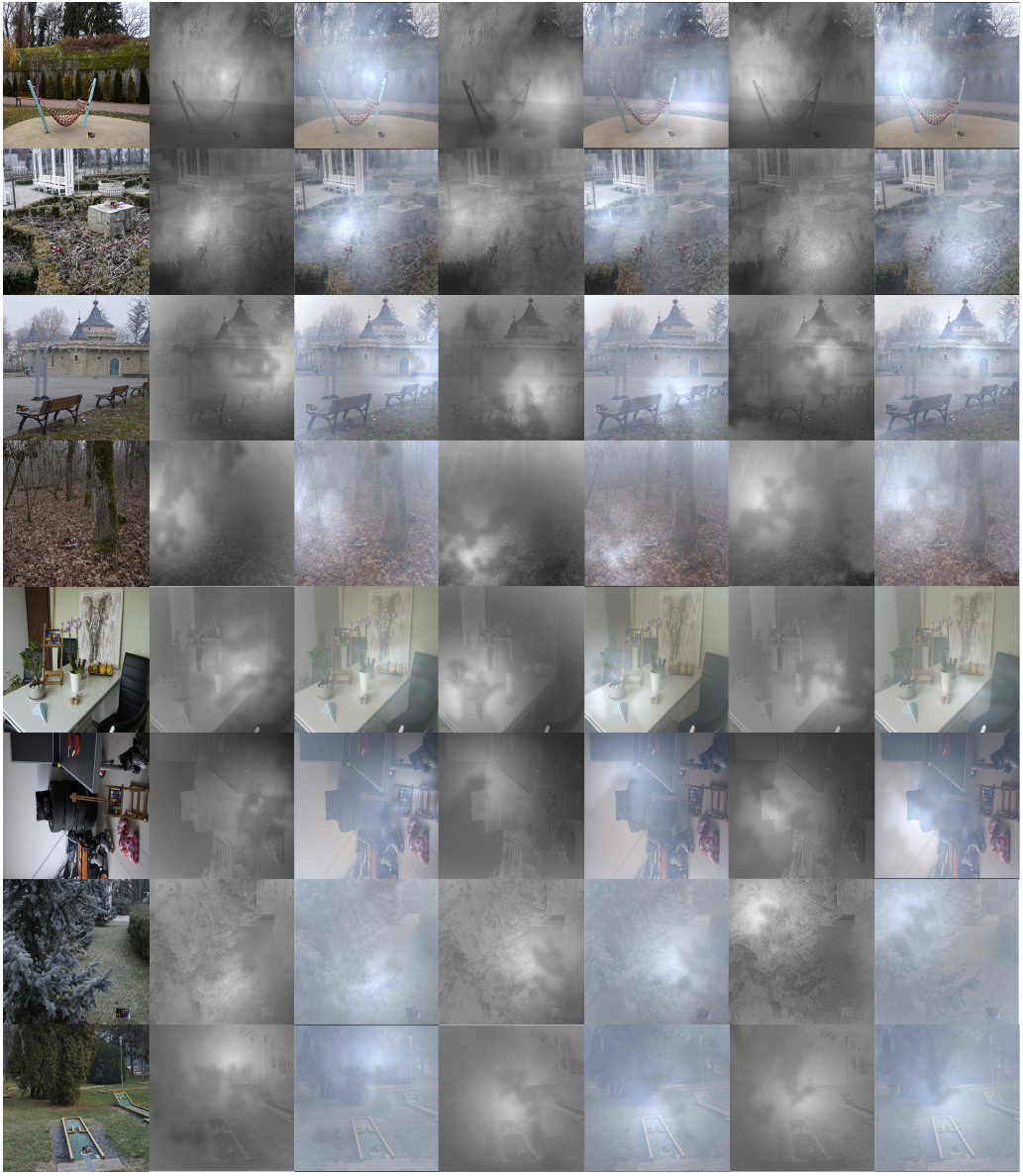} 
\caption{
Additional haze generation results on four paired datasets, showing density maps and corresponding hazy images generated using Equal 5. By applying various random sampling techniques, our approach simulates diverse haze distributions and intensities, closely matching real-world hazy conditions. Each dataset includes two image sets, demonstrating the effectiveness of our method in expanding the training dataset and enhancing dehazing performance on diverse and complex haze distributions.
}
\label{AHG2}
\end{figure*}

\paragraph{AHG-Generated Hazy Images.}  
As shown in Figures~\ref{AHG1} and~\ref{AHG2}, we present additional examples of hazy images generated using our Automatic Haze Generation (AHG) module, along with their corresponding resampled haze density maps.  
These visualizations further demonstrate the ability of AHG to simulate diverse and realistic haze distributions.

\paragraph{More Ablation Study on loss function}
As shown in Table~\ref{tab:ablation_MNCD}, we further investigate the effect of the weight $\lambda_1$ and $\lambda_2$ in our loss function.  

\begin{table}
\centering
\caption{Ablation study on the $\lambda_1$ and $\lambda_2$ in MNCD Loss}
\label{tab:ablation_MNCD}
\resizebox{1.0\linewidth}{!}{
\begin{tabular}{c|ccc}
\hline
$\lambda_1$ & 5 & 1 & 0.1  \\
\hline
AVG PSNR$\uparrow$ & 20.720 & \textbf{21.488} & \underline{21.153} \\

AVG SSIM$\uparrow$ & 0.824 & \textbf{0.849} & \underline{0.837}  \\
\hline
\hline
$\lambda_2$ & 3 & 0.5 & 0.05 \\
\hline
AVG PSNR$\uparrow$ & 20.814 & \textbf{21.488} & \underline{21.123}  \\
AVG SSIM$\uparrow$ & \underline{0.829} & \textbf{0.849} & 0.809  \\
\hline
\end{tabular}}
\end{table}

\end{document}